\documentclass[11pt]{article}

\usepackage[preprint]{acl}

\usepackage{times}
\usepackage{latexsym}
\usepackage[most]{tcolorbox}
\usepackage{booktabs}
\usepackage{makecell}
\usepackage{tabularx}
\usepackage{tikz}
\usetikzlibrary{arrows.meta}
\usepackage[T1]{fontenc}
\usepackage[utf8]{inputenc}
\usepackage{microtype}
\usepackage{inconsolata}
\usepackage{graphicx}
\usepackage{xcolor}

\definecolor{darkgreen}{HTML}{225533}

\title{Reinforcement Learning Restores Reasoning after Corrupted SFT}

\author{%
  Anika Misra \qquad Shreyas Vinod \qquad Huteng Dai \\
  University of Michigan \\
  \texttt{\{anmisra, shreyasc, huteng\}@umich.edu}
}

\begin{document}
\maketitle

\begin{abstract}
We show that reinforcement learning (RL) in reasoning tasks can succeed even when the supervised fine-tuning (SFT) data is severely corrupted. In a mathematical reasoning dataset (GSM8K), we systematically corrupt training trajectories across 361 configurations by injecting realistic error types, varying error locations, trace-answer consistency, and corruption ratios up to 100\%. While aggregated SFT-only performance steadily deteriorates under corruption (dropping from 43.61\% to 13.80\%), Group Relative Policy Optimization (GRPO) post-training consistently restores model accuracy back to gold-baseline levels ($\approx 60\%$). Crucially, initializing RL with as few as $N=8$ minimal,  incorrect, three-line ``form-only'' traces boosts post-RL performance to 69.30\%, outperforming standard gold SFT initialization (60.71\%) and GRPO without SFT (35.00\%). Further, we show that SFT accuracy is largely uncorrelated with post-RL accuracy ($r \approx 0$), demonstrating a fundamental asymmetry in post-training: SFT primarily establishes structural layout (\emph{form}), while RL independently discovers functional reasoning (\emph{function}).
\end{abstract}

\section{Introduction}
\label{sec:intro}

In recent years, Reinforcement Learning (RL) has played a substantial role in advancing Large Language Model (LLM) reasoning capabilities by utilizing reward signals to explore reasoning strategies \cite{ICLR2026_2e53c02e}. To understand the strength of this exploration, our work investigates how well RL can recover from corrupted reasoning. 


Prior works have found success in applying the RL algorithm Group Relative Policy Optimization (GRPO) to enhance mathematical reasoning. Because applying GRPO directly to a base language model with no post-training has shown mixed results in reasoning coherency and stability, supervised fine-tuning (SFT) is traditionally applied as the post-training mechanism beforehand. This is known as the ``SFT-then-RL'' paradigm \cite{shao2024deepseekmathpushinglimitsmathematical, Guo_2025, rajani2025scalpelvshammergrpo, wang-etal-2026-learning-staying}. 

Before applying GRPO, it is standard to optimize for strong SFT performance, particularly selecting high-quality, correct reasoning traces for training \cite{Guo_2025}. However, recent literature challenges the necessity of this curation in two ways. First, recent works have found that strong SFT performance does not necessarily indicate strong post-RL performance \cite{ICLR2026_5d6ae8ba}. Second, it is understood that SFT and RL contribute differing strengths to the learning process \cite{chu2025sftmemorizesrlgeneralizes}. Observing this functional difference and performance non-dependency between SFT and RL motivates our first question: 
\begin{figure}[t]
  \centering
  \definecolor{fgInk}{HTML}{3F3F46}
  \definecolor{fgModel}{HTML}{3D6FA3}
  \definecolor{fgGold}{HTML}{E6B33E}
  \definecolor{fgGoldD}{HTML}{A8790F}
  \definecolor{fgErr}{HTML}{D1495B}
  \definecolor{fgOk}{HTML}{2A9D8F}
  \begin{tikzpicture}[
      x=1.22cm, y=1.22cm,
      font=\sffamily\scriptsize,
      line cap=round, line join=round,
      >={Stealth[length=3.4pt,width=3pt]},
      flow/.style={->, line width=.6pt},
      hop/.style={-{Stealth[length=2.6pt,width=2.4pt]}, fgErr, line width=.5pt},
      lbl/.style={text=fgInk, inner sep=1pt, anchor=south},
    ]
    \begin{scope}[shift={(.35,0)}]
      \fill[fgInk!7] (-.27,.28) -- (-.27,-.28)
        arc[start angle=180, end angle=360, x radius=.27, y radius=.08] -- (.27,.28) -- cycle;
      \foreach \yy in {.09,-.10}
        \draw[fgInk!40, line width=.4pt] (-.27,\yy)
          arc[start angle=180, end angle=360, x radius=.27, y radius=.08];
      \draw[fgInk!75, line width=.5pt] (-.27,.28) -- (-.27,-.28)
        arc[start angle=180, end angle=360, x radius=.27, y radius=.08] -- (.27,.28);
      \draw[fgInk!75, fill=fgInk!18, line width=.5pt] (0,.28) ellipse[x radius=.27, y radius=.08];
    \end{scope}
    \node[lbl] at (.35,.42) {GSM8K};

    \begin{scope}[shift={(1.85,0)}]
      \draw[fgModel!50, line width=.5pt, rounded corners=2.5pt,
            top color=fgModel!14, bottom color=fgModel!3] (-.52,-.40) rectangle (.52,.40);
      \foreach \a in {-1,0,1} \foreach \b in {-1.5,-.5,.5,1.5}
        \draw[fgModel!45, line width=.3pt] (-.32,\a*.21) -- (0,\b*.19) -- (.32,\a*.21);
      \foreach \a in {-1,0,1} {
        \fill[fgModel] (-.32,\a*.21) circle (.05);
        \fill[fgModel] ( .32,\a*.21) circle (.05);
      }
      \foreach \b in {-1.5,-.5,.5,1.5} \fill[fgModel] (0,\b*.19) circle (.05);
    \end{scope}
    \node[lbl] at (1.85,.42) {OLMo-2-1B};
    \draw[flow, fgInk] (.68,0) -- (1.29,0);

    \draw[fgModel!60, line width=.45pt] (2.37,0) -- (2.56,0);
    \foreach \yy in {.285,.095,-.095,-.285} {
      \draw[fgModel!60, line width=.45pt] (2.56,0) to[out=0, in=180] (3.08,\yy);
      \foreach \k in {0,1,2}
        \fill[fgModel!38, rounded corners=.5pt] (3.14+\k*.155,\yy-.06) rectangle ++(.12,.12);
    }
    \foreach \yy in {.285,-.095} {
      \fill[fgOk] (3.69,\yy) circle (.076);
      \draw[white, line width=.6pt] (3.655,\yy) -- (3.68,\yy-.026) -- (3.725,\yy+.028);
    }
    \foreach \yy in {.095,-.285} {
      \fill[fgErr] (3.69,\yy) circle (.076);
      \draw[white, line width=.6pt] (3.663,\yy-.027) -- (3.717,\yy+.027)
                                    (3.663,\yy+.027) -- (3.717,\yy-.027);
    }
    \node[lbl] at (3.40,.42) {GRPO};

    \foreach \yy in {.095,-.285}
      \draw[fgErr, line width=.5pt] (3.76,\yy) to[out=0, in=180] (4.36,0);
    \draw[flow, fgErr] (4.36,0) -- (4.84,0);

    \begin{scope}[shift={(5.12,0)}]
      \fill[fgInk!38] (-.10,.15) circle (.075);
      \fill[fgInk!38] (-.23,-.17) .. controls (-.23,.06) and (.03,.06) .. (.03,-.17) -- cycle;
      \filldraw[fgInk!80, draw=white, line width=.7pt] (.08,.10) circle (.08);
      \filldraw[fgInk!80, draw=white, line width=.7pt]
        (-.07,-.22) .. controls (-.07,.02) and (.23,.02) .. (.23,-.22) -- cycle;
    \end{scope}
    \draw[flow, fgInk] (5.12,-.29) -- (5.12,-.74);

    \fill[fgInk!5, draw=fgInk!25, line width=.4pt, rounded corners=2pt]
      (3.24,-1.15) rectangle (5.31,-.75);
    \foreach \i in {1,...,5}
      \draw[fgInk!30, fill=white, line width=.4pt, rounded corners=1pt]
        (3.31+\i*.33,-1.09) rectangle ++(.28,.28);
    \fill[fgErr, rounded corners=1pt] (3.31,-1.09) rectangle ++(.28,.28);
    \fill[white] (3.45,-.865) -- (3.37,-1.01) -- (3.53,-1.01) -- cycle;
    \begin{scope}[fgInk!70]
      \fill (3.78,-.95) circle (.07);
      \fill (4.048,-1.012) rectangle ++(.124,.124);
      \fill (4.44,-.86) -- (4.515,-.95) -- (4.44,-1.04) -- (4.365,-.95) -- cycle;
      \fill[shift={(4.77,-.955)}] (90:.085) -- (162:.085) -- (234:.085) -- (306:.085) -- (18:.085) -- cycle;
      \fill[shift={(5.10,-.955)}] (90:.095) -- (126:.04) -- (162:.095) -- (198:.04) -- (234:.095)
        -- (270:.04) -- (306:.095) -- (342:.04) -- (18:.095) -- (54:.04) -- cycle;
    \end{scope}
    \draw[flow, fgErr] (3.45,-1.15) -- (3.45,-1.61);
    \node[text=fgInk, anchor=east, inner sep=1pt] at (3.39,-1.37) {regex};

    \draw[flow, fgGoldD] (.35,-.40) |- (1.27,-1.80);
    \foreach \x in {1.45,1.95,2.45,2.95}
      \draw[fgGoldD, fill=fgGold!60, line width=.5pt, rounded corners=1pt]
        (\x-.16,-1.96) rectangle ++(.32,.32);
    \draw[fgErr!80!black, fill=fgErr, line width=.5pt, rounded corners=1pt]
      (3.29,-1.96) rectangle ++(.32,.32);
    \fill[white] (3.45,-1.70) -- (3.365,-1.86) -- (3.535,-1.86) -- cycle;
    \foreach \x/\t in {3.95/22, 4.45/36, 4.95/50}
      \draw[fgErr!75, fill=fgErr!\t, line width=.5pt, rounded corners=1pt]
        (\x-.16,-1.96) rectangle ++(.32,.32);
    \fill[fgErr] (5.52,-1.80) circle (.17);
    \draw[white, line width=1pt] (5.46,-1.86) -- (5.58,-1.74) (5.46,-1.74) -- (5.58,-1.86);
    \foreach \a/\b in {3.45/3.95, 3.95/4.45, 4.45/4.95}
      \draw[hop] (\a,-2.00) to[bend right=60] (\b,-2.00);
    \draw[hop] (4.95,-2.00) to[bend right=50] (5.45,-1.97);
  \end{tikzpicture}
  \caption{Pipeline overview. Two human judges sort incorrect GRPO rollouts from OLMo-2-1B into six error types (symbols). Regex injects one type into a step of a gold GSM8K trace (squares) and propagates it to the final answer (circle).}
  \label{fig:pipeline}
\end{figure}
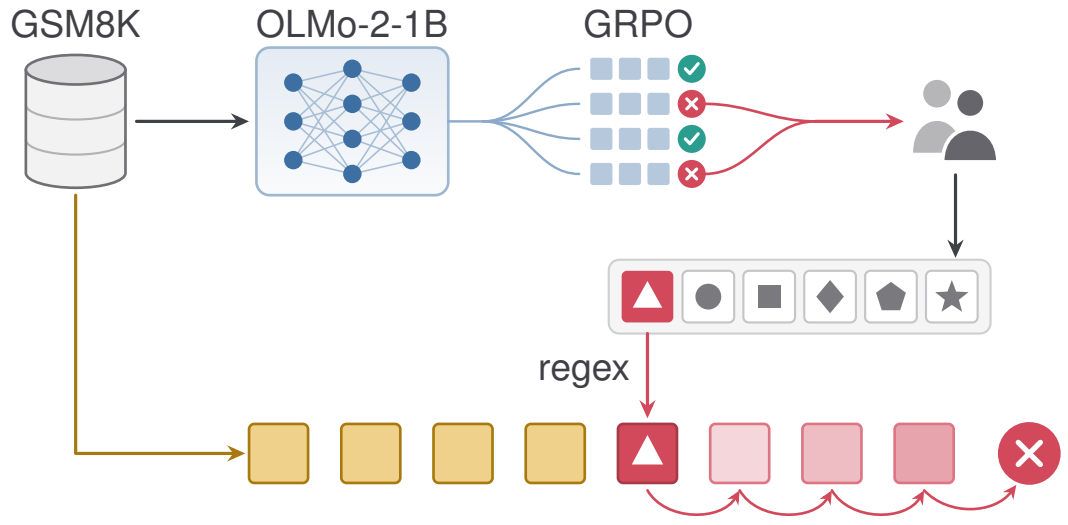
\textit{(1) How necessary is high-quality data in the SFT stage prior to GRPO?}

The data filtering method of previous approaches removes reasoning traces that result in incorrect final answers among other data quality issues \citep[e.g., DeepSeek R1; ][]{Guo_2025}. Following this, we conduct a diagnostic GRPO run, filter incorrect traces, and manually identify common error types. We then build a corrupted dataset by simulating and propagating these errors in math reasoning traces using regex. 
Finally, we conduct SFT on each training configuration before applying GRPO.

Our results display remarkably powerful improvement capabilities by GRPO. Across all training configurations, SFT models steadily deteriorate as the percentage of corrupted data increases. However, at every single configuration, GRPO corrects performance to near-gold baseline levels, up to 100\% corruption (Figure \ref{fig:grpo_vs_sft}). GRPO's insensitivity to correct SFT reasoning leads us to our second question: \textit{(2) Can RL be successfully initialized with form alone?} 

To answer this question, we construct bare-bones, minimal trace data with just three lines in each trace, denoted as \textbf{form-only traces}. We conduct SFT on as few as $N=8$ samples before applying GRPO once more. 

Strikingly, models trained with form-only SFT achieve even higher post-RL performance than models trained with standard, gold SFT. 
These patterns hold consistent whether the form-only traces contain \textit{correct} arithmetic reasoning or not. Notably, an SFT stage prior to GRPO is always beneficial, even in extreme corruption scenarios.

Overall, these results suggest that SFT teaches the model something imperative, but not necessarily reasoning. GRPO remains robust as SFT performance deteriorates under increasingly corrupted supervision, and displays even stronger performance when SFT data is reduced to minimal, form-only traces. This indicates that post-RL performance can strongly benefit when SFT and RL are assigned fundamentally different roles in the learning process. We interpret these roles as \textbf{form} taught by SFT and \textbf{function} taught by RL. This separation offers an explanation for the observed non-dependency between SFT and post-RL performance, and suggests that SFT can be optimized for something other than reasoning quality to obtain strong performances in the ``SFT-then-RL'' paradigm. 

\section{Related Work}
\label{sec:related_work}
\textbf{Evolution of Language Models.}
Early neural language models relied on recurrent neural network frameworks such as Long-Short Term Memory to predict word sequences \cite{NIPS2014_5a18e133}. The transformer architecture addressed prior sequential computation limitations, enabling efficient pre-training of large models on vast amounts of data and unlocking the advanced reasoning capabilities we see in LLMs today \cite{NIPS2017_3f5ee243, Zhao2026}. 

\textbf{Reasoning in LLMs.} 
To elicit reasoning in LLMs after pre-training, supervised fine-tuning (SFT) on high-quality reasoning traces, prompting methods such as Chain-of-Thought, and reinforcement learning (RL) post-training methods may all be applied \cite{JMLR:v25:23-0870, magister-etal-2023-teaching, chu2025sftmemorizesrlgeneralizes}. SFT learns by imitating demonstrated reasoning, while RL utilizes reward signals to explore reasoning strategies \cite{matsutani2026rl}. Initial RL methods for LLMs became important in human preference alignment, relying on an external reward model trained on high-quality preference data \cite{NEURIPS2022_b1efde53}. Conversely, reasoning tasks with verifiable outcomes such as math or coding do not require a trained reward model and can rely solely on answer correctness \cite{Guo_2025, wen2025reinforcementlearningverifiablerewards}. Within verifiable tasks, Group Relative Policy Optimization (GRPO) normalizes rule-based rewards across a group of sampled outputs, bypassing the necessity of a critic model \cite{shao2024deepseekmathpushinglimitsmathematical}. 

\textbf{Performance Decoupling in SFT and RL.}  SFT is utilized as a cold-start initialization to format model responses before applying RL \cite{Guo_2025}. Though standard alignment procedures optimize both the SFT and the RL phase, recent works have shown that high-performing SFT does not always indicate high-performing post-RL \cite{ICLR2026_5d6ae8ba}. Excessive SFT can teach the model memorization and limit the exploration process of RL \cite{chu2025sftmemorizesrlgeneralizes}. Our work formalizes this decoupling by targeting data corruption and simple trace structure in the SFT process. 

\textbf{Corrupting Reasoning Traces.} Prior works have examined effects on LLM reasoning capabilities after corrupting SFT data, finding that corrupting function, not form, leads to SFT-model degradation \cite{cho-2024-unveiling, li-etal-2025-language-models}. We build on these works in two ways: (1) we ground our errors in real, model-generated errors rather than synthetic ones, and (2) we move beyond the SFT stage, analyzing instead the effects of these realistic errors as they are improved upon with RL. 

\section{Methods}
\label{sec:methods}
\begin{figure*}[t]
    \centering
    \includegraphics[width=0.8\textwidth]{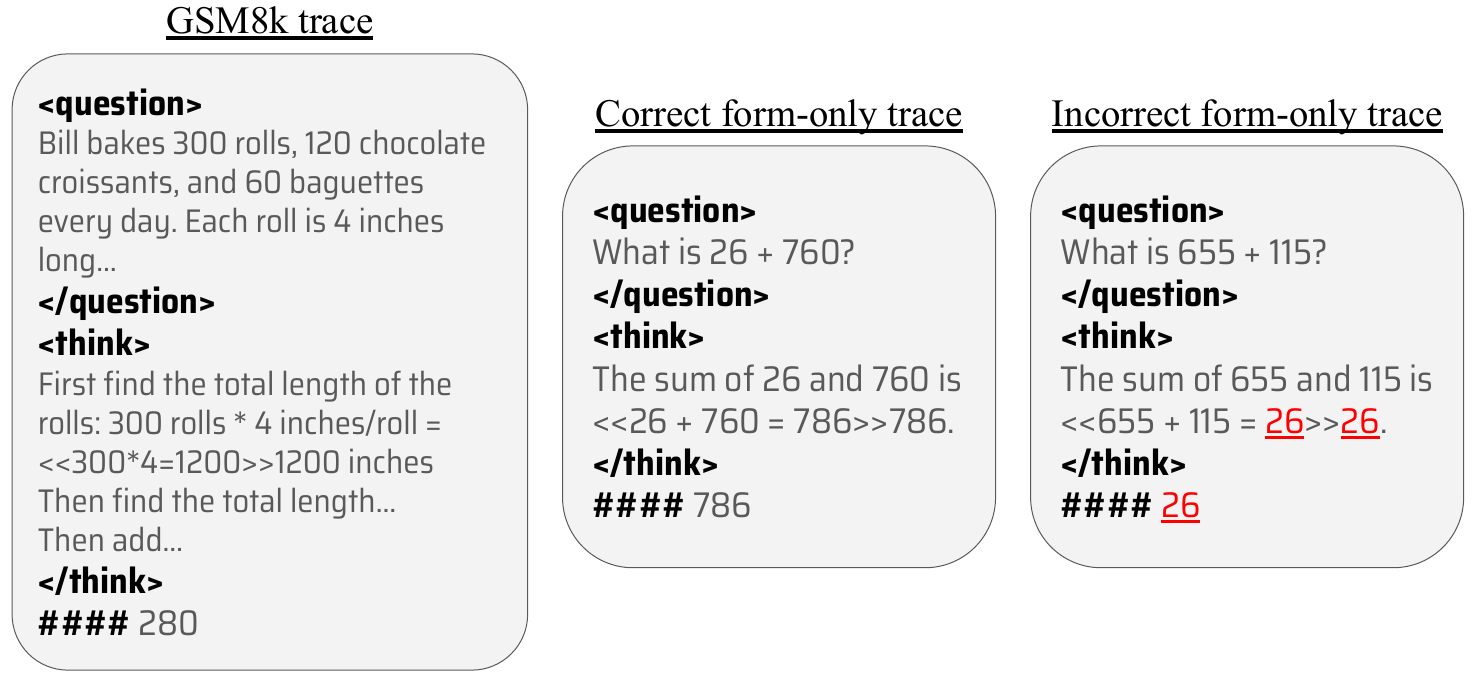}
    \caption{GSM8K samples consist of diverse, multi-step mathematical reasoning. Form-only traces consist of solely the basic trace structure, with a correct and incorrect variant.}
    \label{fig:form_only_data_examples}
\end{figure*}
\subsection{Model and dataset}

To establish a controlled diagnostic testbed, we intentionally restrict our experiments to the OLMo-2-1B base model with the GSM8K English math reasoning dataset \cite{olmo20242olmo2furious, cobbe2021trainingverifierssolvemath}. GSM8K's simple format and validated reasoning traces enable the use of a deterministic, regex-based corrupter, and the 1B model size enables a systematic sweep of a broad corruption space while still training each condition to performance saturation. The 1B model additionally provides an unsaturated starting point (21.99\% pass@$1$ accuracy) for analyzing intervention effects. Finally, OLMo-2 provides publicly available training data and explicit tests of test data leakage, limiting contamination from prior task knowledge \cite{olmo20242olmo2furious}. 

\subsection{Initial GRPO run}
Despite scoring highly on answer-only evaluations, models trained to produce reasoning traces with no SFT prior to GRPO show mixed effectiveness at producing comprehensible, accurate reasoning traces \cite{Guo_2025}. We conduct a diagnostic GRPO run with minimal SFT on a base non-thinking model to collect traces and explore real, common failure modes in incorrect traces.

From this initial experiment, we identify two key findings of interest. First, incorrect traces generally fall into categorizable groups based on the specific error in the trace. Second, errors early in traces typically propagate through to the end.

\subsection{Error analysis}
After conducting the initial GRPO run, we filtered 601 test traces that resulted in an incorrect final answer. We manually inspected thirty incorrect traces to establish six common error types. Two authors then utilized a shared rubric to categorize the resulting incorrect traces into six error types, and edge cases were resolved with mutual consensus. 93.51\% of incorrect traces fall into one or more of the defined error types. The choice of human inspection was to reduce the risk of circularity or undetected errors overlapping with a judge model. Section \ref{sec:corrupter} details these error types and provides examples.

\subsection{Regex-based propagation}
\begin{figure*}[t]
    \centering
    \includegraphics[width=\textwidth]{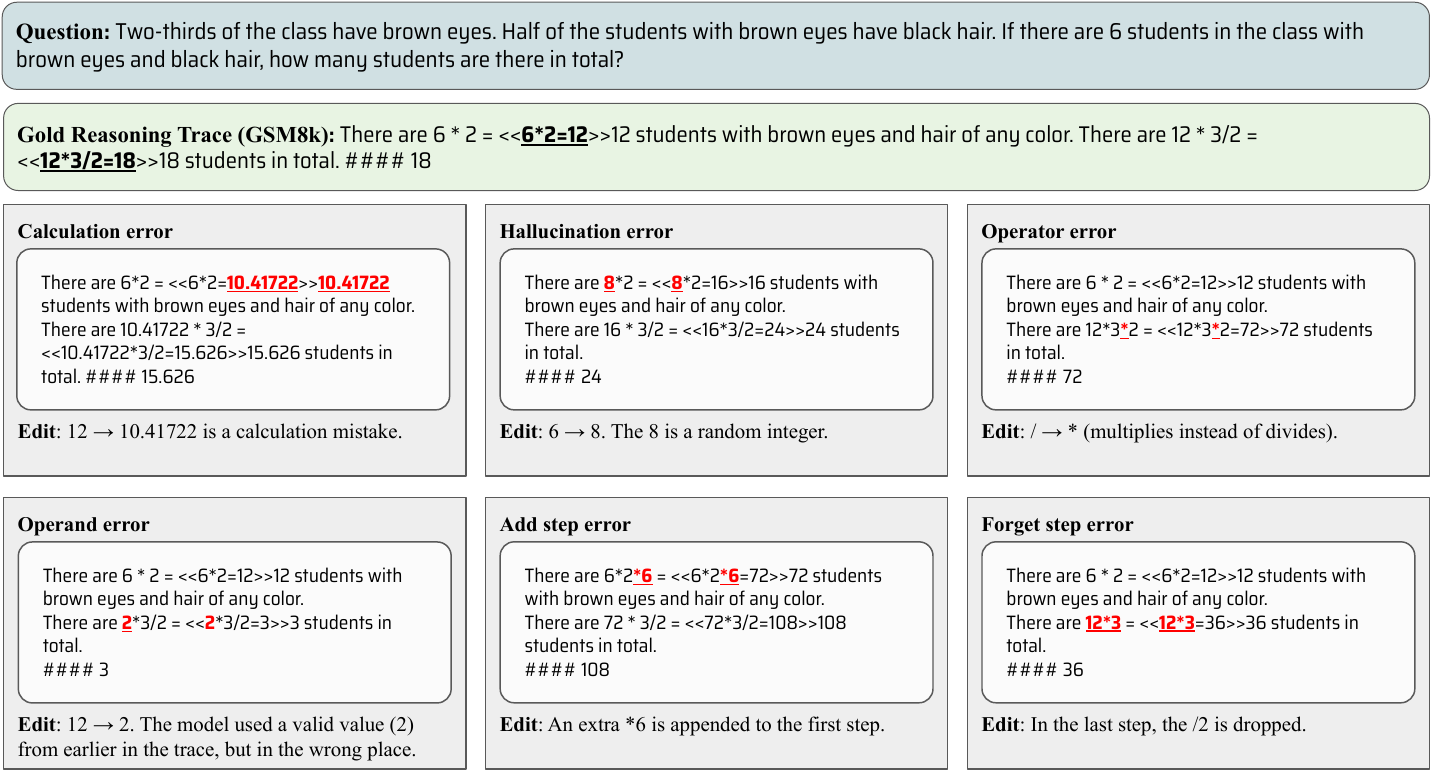}
    \caption{Example of our six corruption types injected into a GSM8K reasoning trace. For each corruption type, we use regex to modify a value and deterministically propagate the error through, resulting in an incorrect final answer.}
    \label{fig:corruption_examples}
\end{figure*}

To corrupt a reasoning trace with a specific error type, we use regex to select an equation in the trace, apply the corruption, and propagate the error through the rest of the trace (Figure \ref{fig:corruption_examples}). We avoid LLM-based propagation to ensure strict control over the computations and downstream propagation, as prior work finds that LLMs are prone to arithmetic hallucinations \cite{yuan2023largelanguagemodelsperform}. 

Traces where steps were skipped or implied were unsuccessful with regex-based propagation because mathematical results did not directly appear in subsequent equations. Starting from 7473 GSM8K samples, we obtain 6913 regex-corruptible traces to be used as training data in our experiments.

\subsection{Form-only traces}
In addition to training on GSM8K traces, we construct minimal, three-line form-only traces. These represent basic trace structure rather than detailed step-by-step mathematical reasoning. Specifically, we construct two types. \textbf{Correct form-only} traces consist of two randomly sampled integers with an arithmetically correct sum, while \textbf{corrupted form-only} traces consist of three randomly sampled integers arranged in an addition-like expression, without true arithmetic correctness. Examples are shown in Figure \ref{fig:form_only_data_examples}.

\subsection{Corrupter}
\label{sec:corrupter}

After conducting our diagnostic GRPO run, we identify and implement six corruption types as described below: 
\begin{enumerate}
    \item \textbf{Calculation error}: the model makes an arithmetic mistake. 
    \item \textbf{Hallucination error}: the model introduces a random number in the trace. 
    \item \textbf{Operator error}: the model uses the incorrect operator for the task (e.g. multiplying instead of dividing).
    \item \textbf{Operand error}: the model uses a valid value from the trace or problem statement in the wrong place.
    \item \textbf{Add step error}: the model appends an unnecessary step to its calculations. 
    \item \textbf{Forget step error}: the model forgets to perform a critical step.
\end{enumerate}

To simulate each error type, we modify the appropriate value in a gold GSM8K reasoning trace, recalculate the result of the equation where applicable, update the equation, and propagate this new value through to result in an incorrect final answer. Examples of each corruption type are shown in Figure \ref{fig:corruption_examples}. Further corruption and propagation details can be found in Appendix \ref{appendix:corrupter_details}. 

\section{Experiments}
\label{sec:experiments}

\subsection{Corrupted training data configurations}
\label{sec:experiments:training_data_configs}

To systematically evaluate the effect of training data corruption on our post-training methods, we vary the corruption ratio from 10\% to 100\% in increments of 10\%. Because SFT learns from both the final answer and the reasoning trajectory, we additionally decouple answer correctness from trace correctness in the tuple $(trace, answer) \in \{0,1\}^2$: 
\begin{itemize}
    \item \textbf{(0,0)}: both the trace and the final answer are incorrect, as shown in Figure \ref{fig:corruption_examples}.
    \item \textbf{(0,1)}: the trace is incorrect, but the final answer is swapped for the gold answer.
    \item \textbf{(1,0)}: the trace is correct, but the final answer is swapped for an incorrect one.
\end{itemize}   

To analyze the effect of error depth, we separately corrupt the first and last extracted equation in the trace, propagating the error forward as needed. We finally apply the 6 different corruption types described in Section \ref{sec:corrupter}. With 6 error types, 2 locations, 3 trace-answer configurations, 10 corruption ratios, and 1 gold baseline, we obtain $\{6 \times 2 \times 3 \times 10 \} + 1 = 361$ SFT-model configurations.

\subsection{GRPO-only}
\label{sec:experiments:grpo-only}
To investigate how reasoning abilities may emerge on their own, we began with a GRPO run on the full, uncorrupted GSM8K dataset (7473 samples), omitting the SFT-pretraining step entirely. We trained for 40k steps with 8 rollouts per step, $3\times10^{-7}$ learning rate, constant-learning rate schedule with 3\% warmup, and $\beta$=0.01 (hyperparameter details in Appendix \ref{appendix:hyperparameters}). The resulting traces were poorly formatted, necessitating some minimal SFT to teach the model trace structure before analyzing model reasoning failures.  

\subsection{Diagnostic GRPO}
\label{sec:experiments:grpo_with_arith_sft}
To teach the model basic trace structure, we construct 50 form-only samples and conduct SFT for 4 epochs with a learning rate of $2\times10^{-5}$, and a cosine learning-rate schedule with a 10\% warmup (hyperparameter details in Appendix \ref{appendix:hyperparameters}). To isolate RL-generated reasoning failures, we required that the model had attained trace coherency but had not yet converged on mathematical correctness. We use the 5,000-step checkpoint to obtain structurally coherent but logically flawed reasoning traces, which we then analyze to understand common model failure modes as described in Section \ref{sec:methods}.

\subsection{SFT-only}
\label{sec:experiments:sft-only}
After identifying common model error types to build our corrupted dataset, we conduct SFT on each of the 361 training data configurations with the same training parameters as Section \ref{sec:experiments:grpo_with_arith_sft}. We specifically analyze the SFT-only results before applying GRPO to separately understand the effects of incorrect reasoning data on each learning dynamic. 

\begin{figure}[t]
    \centering
    \includegraphics[width=\columnwidth]{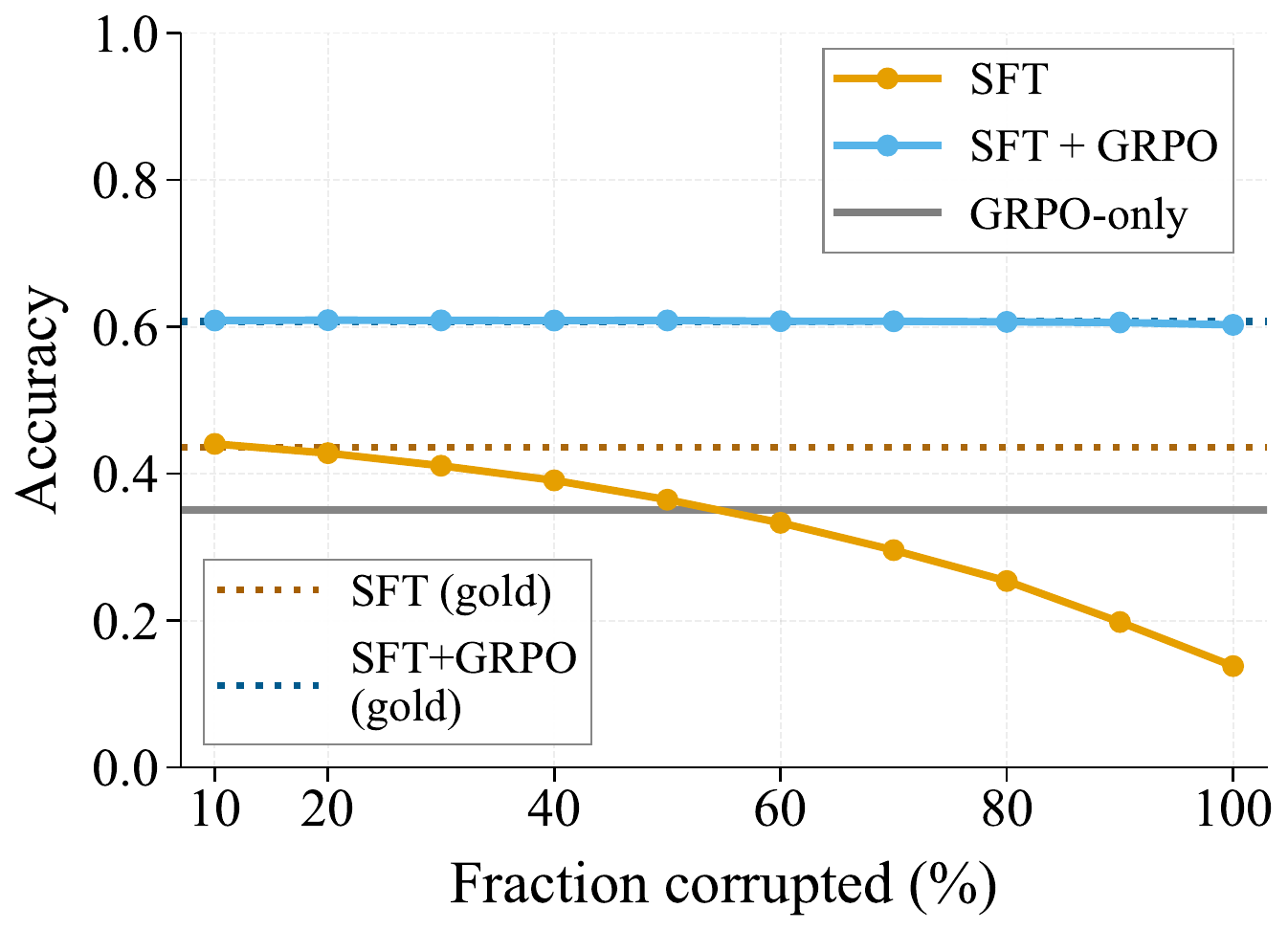}
    \caption{Aggregated across all corruption configurations, as the fraction of corrupted training data increases, SFT model accuracy deteriorates, but GRPO post-training corrects this behavior and restores accuracy to near-gold baselines, even up to 100\% corruption.}
    \label{fig:grpo_vs_sft}
\end{figure}

\subsection{SFT + GRPO}
Next, we apply GRPO to each corresponding SFT-model configuration. This follows state-of-the-art methods of SFT to establish stronger task performance before reinforcement learning \cite{matsutani2026rl}. We use the same GRPO training parameters as in Section \ref{sec:experiments:grpo-only}. 

\subsection{Form-only}
To analyze how RL can be initialized from minimalistic trace structure, we conduct SFT on form-only datasets of size $N \in \{8, 16, 32, 64, 128, 256\}$. Finally, on each SFT-trained model, we conduct GRPO using the full, uncorrupted GSM8K dataset. 

\subsection{Evaluation}
All final accuracies are computed on the GSM8K test set (1319 samples). Our primary evaluation metric is pass@$2$, which is reported for all main findings unless otherwise specified. Further evaluation details and pass@$k$ results are in Appendix \ref{appendix:evaluation_details} and \ref{appendix:other_pass_@_k_results}. 

\section{Results}
\label{sec:results}

\subsection{SFT degrades, but RL corrects}

In our GSM8K-trained SFT configurations, increasing the ratio of corrupted training data induces a steady decline in aggregated SFT-only model accuracy, dipping from 43.61\% accuracy at the gold SFT-only baseline to 13.80\% accuracy at 100\% corrupted ratio. Yet, at every corruption ratio, post-GRPO consistently restores accuracy to gold-baseline levels of approximately 60\% in the SFT+GRPO configuration (Figure \ref{fig:grpo_vs_sft}). While the highest aggregated SFT data corruption ratios result in minor post-GRPO performance dips, the maximum observed aggregated deviation from the gold SFT+GRPO baseline never exceeds 0.5 percentage points. These recovery patterns extend to our form-only training configurations. SFT on corrupted form-only data results in worse performance than regular form-only data by up to 16.51 percentage points. Still, post-GRPO consistently restores model performance to similar levels for both regimes. In fact, at sample sizes of $N=8$ and $N=64$, post-GRPO on corrupted form-only samples outperforms post-GRPO on regular form-only samples.

\subsection{GRPO is more robust to corrupted initializations}

\begin{figure}[t]
    \centering
    \includegraphics[width=\columnwidth]{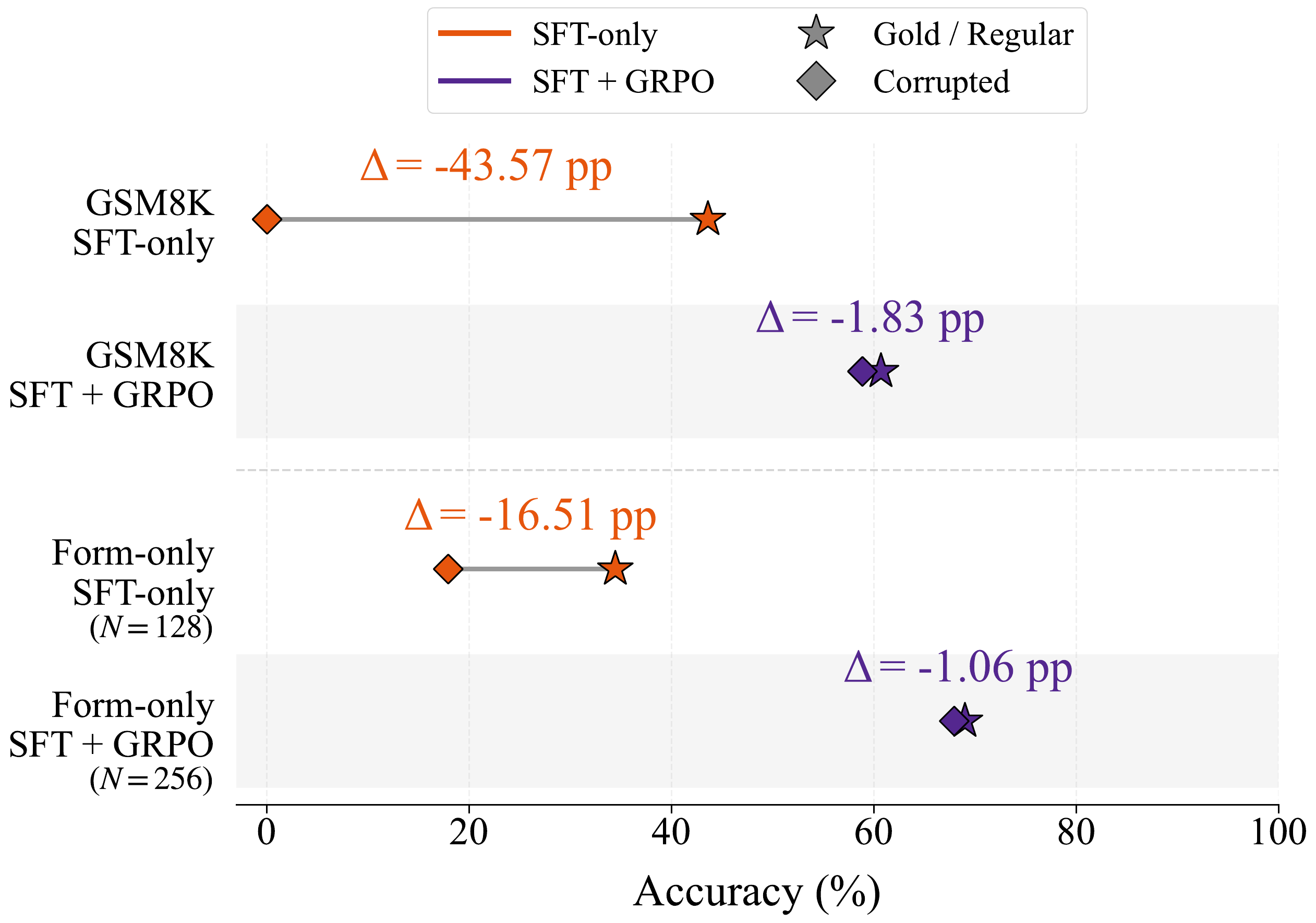}
    \caption{GRPO displays robustness to corrupted SFT initialization. The absolute accuracy differences between the lowest-performing corrupted configuration and its gold-initialized counterpart are larger in SFT-only (orange) than in SFT+GRPO (purple).}
    \label{fig:worst_corruption}
\end{figure}

Across our experiments, we observe GRPO robustness to corrupted SFT initializations, displayed in Figure \ref{fig:worst_corruption}. 
The worst-performing corrupted GSM8K initialization exhibits a 43.57 percentage-point decline in SFT-only accuracy relative to the gold initialization. In contrast, the worst-performing corrupted initialization after GRPO exhibits a decline of only 1.83 percentage points relative to the gold SFT+GRPO baseline. Similarly, the worst-performing corrupted form-only initialization exhibits a 16.51 percentage-point decline in SFT-only accuracy relative to its regular form-only counterpart, while the largest decline in SFT+GRPO accuracy is 1.06 percentage points.

\subsection{Form-only initialization outperforms standard initialization}
\begin{figure}[t]
    \centering
    \includegraphics[width=\columnwidth]{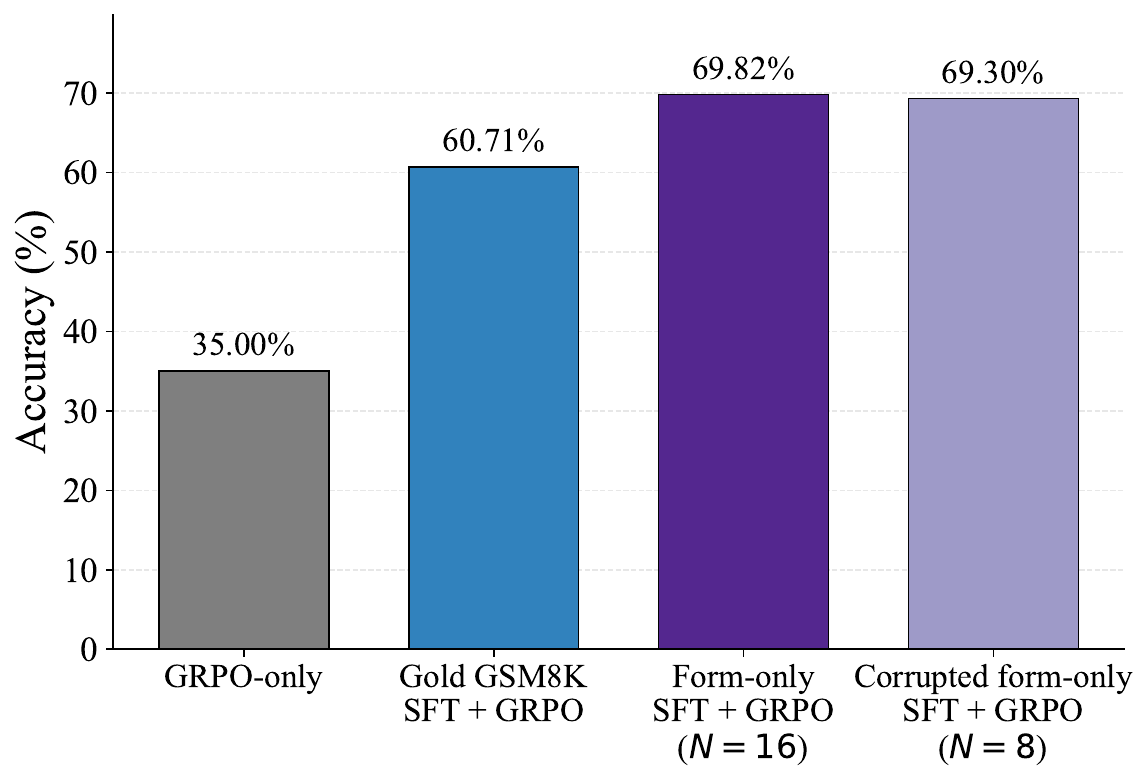}
    \caption{Gold GSM8K SFT + GRPO outperforms the GRPO-only baseline, while minimal form-only initializations achieve the highest overall accuracy.}
    \label{fig:sft_gsm8k_vs_form_only_gsm8k}
\end{figure}

Initializing GRPO with small sets of form-only traces improves performance not only over the GRPO-only baseline, but over the standard gold GSM8K initialization as well. As shown in Figure \ref{fig:sft_gsm8k_vs_form_only_gsm8k}, the gold GSM8K-trained baseline achieves an accuracy of 60.71\% after applying GRPO. In contrast, both regular and corrupted form-only initializations surpass this result, outperforming the gold baseline by up to 9.11 percentage points and 8.59 percentage points, respectively. This performance advantage holds constant: with a single exception of the $N=8$ correct form-only initialization (60.93\% post-GRPO accuracy), nearly all form-only SFT initializations outperform each of the 361 GSM8K SFT initializations.

\subsection{SFT accuracy is weakly associated with post-GRPO performance}
\begin{figure*}[t]
    \centering
    \includegraphics[width=0.8\textwidth]{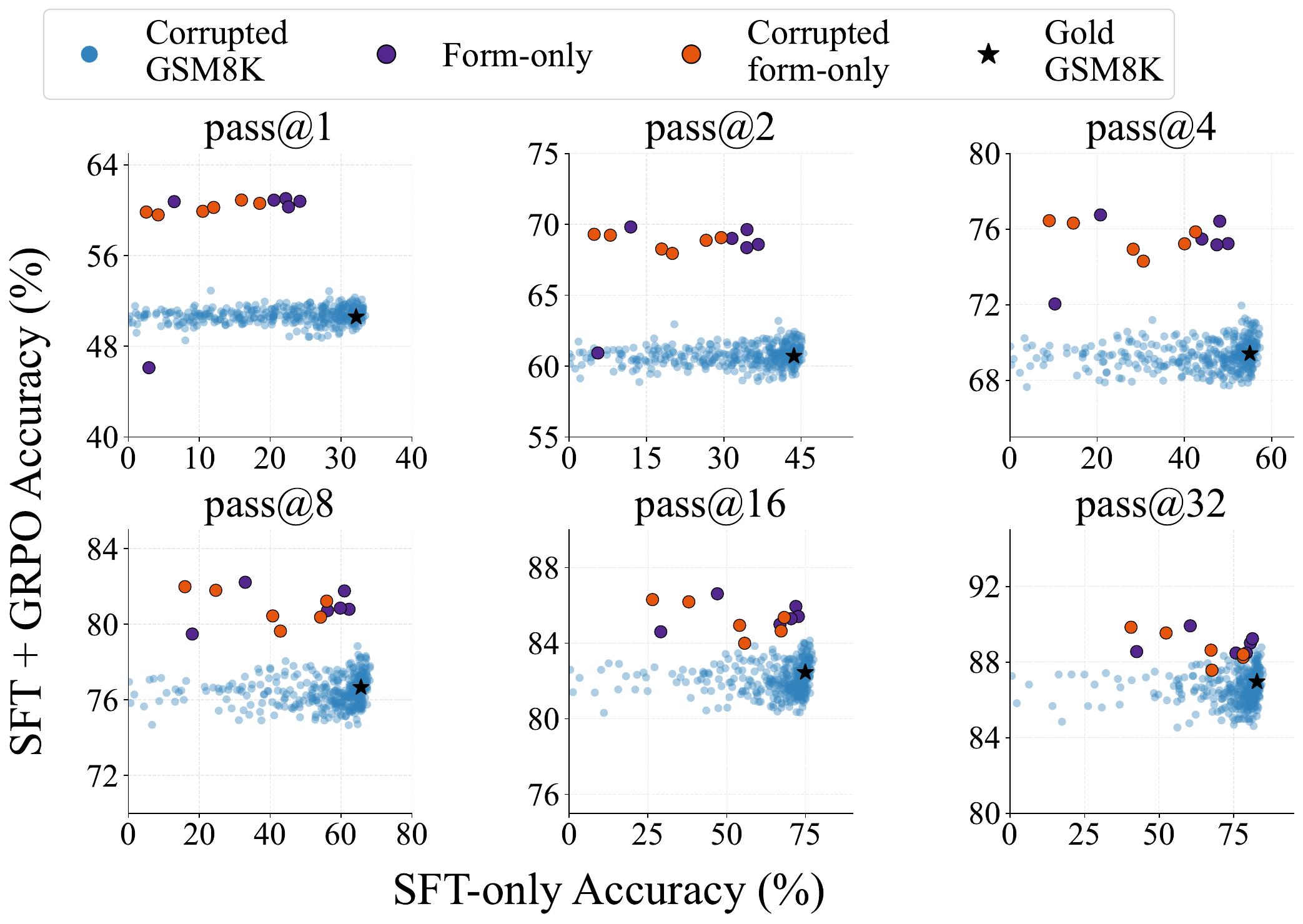}
    \caption{When analyzing data points from all training configurations across each pass@$k$ value, we do not observe a clear visual association between higher SFT-only accuracy and higher post-GRPO accuracy.}
    \label{fig:correlations}
\end{figure*}

To formalize the separation of SFT initialization quality from final GRPO outcomes, we analyze the correlation between SFT-only accuracy and SFT+GRPO accuracy. To control for higher pass@$k$ values reflecting increasingly higher chances of correct answers, we separate each pass@$k$ value \cite{dragoi2025passkbreadthdepthmetricsreasoning}. Visually, we do not observe a clear linear relationship between SFT-only accuracy and post-GRPO accuracy after plotting in Figure \ref{fig:correlations}. Pearson correlation coefficients, shown in Table \ref{tab:pearson_corr}, across all pass@$k$ values approach zero, ranging from $r=-0.0631$ to $r=0.0002$. None of these correlations are statistically significant ($p>0.05$ for all $k$), suggesting that strong reasoning performance in the SFT stage is not a necessity for successful RL post-training.

\begin{table}[t]
\centering
\resizebox{\columnwidth}{!}{
\begin{tabular}{c|cccccc}
\toprule
Pass@k & 1 & 2 & 4 & 8 & 16 & 32 \\
\midrule
Pearson $r$ & -0.0631 & -0.0569 & -0.0461 & -0.0309 & -0.0055 & 0.0002 \\
\bottomrule
\end{tabular}
}
\caption{Pearson correlations between SFT-only and SFT+GRPO accuracies across all training configurations remain low for each pass@$k$ value.}
\label{tab:pearson_corr}
\end{table}

\subsection{SFT before GRPO is still needed}
Although we do not find a correlation between SFT accuracy and final GRPO performance, some SFT is still beneficial for initialization. Across all evaluated regimes, all forms of SFT prior to GRPO outperformed the GRPO-only baseline (Figure \ref{fig:sft_gsm8k_vs_form_only_gsm8k}). Even conducting SFT on just 8 corrupted form-only samples led to a final post-GRPO accuracy of 69.30\%, up from the GRPO-only baseline of 35.00\%. This suggests that while RL may discover reasoning \textit{function}, it still fundamentally relies on SFT to establish the \textit{form} required.

\section{Discussion}
\label{sec:discussion}


\subsection{Why RL recovers from corrupted SFT initializations}

Across our experiments, we observe GRPO recovery and robustness to low-performing, corrupted SFT initializations. We hypothesize these recovery and robustness abilities arise from the preservation of \textit{form} in corrupted data. Our corrupted traces propagate a model error throughout a trace, but the overall structure of the reasoning trace remains intact (Figure \ref{fig:corruption_examples}). SFT is efficient at imitating expert demonstrations, while RL excels at generalization and reasoning \cite{chu2025sftmemorizesrlgeneralizes}. Hence, our reasoning data, even when corrupted, provides enough structural form to stabilize the model's output format during SFT before utilizing RL to explore and learn the reasoning \textit{function}. 

\subsection{Why minimal-form initialization outperforms full reasoning SFT}
All form-only initializations, including corrupted and regular ones, outperform full gold GSM8K initialization after SFT is applied. Though  both the full GSM8K and minimal form-only traces provide \textit{form} during SFT initialization, there are two main differences between the setups. First, the GSM8K traces contain detailed reasoning trajectories and strategies tailored to the GSM8K problem type, whereas the form-only traces contain single line reasoning for basic arithmetic. Second, form-only SFT trains on far fewer samples than full GSM8K SFT. Prior work finds that expert reasoning patterns learned during SFT guide initial RL exploration, but over-training during SFT can limit this exploration \cite{zhang2026onpolicyrlmeetsoffpolicy, ICLR2026_5d6ae8ba}. We hypothesize that combining detailed, GSM8K-specific reasoning strategies with a higher amount of training leads to more constrained RL exploration in the GSM8K initialization than in the form-only initialization. The constrained reasoning space limits discovery of superior reasoning pathways \cite{zhang2026onpolicyrlmeetsoffpolicy}, hurting model performance. 

\subsection{Generalizability and scale}

The Superficial Alignment Hypothesis proposes that a model's underlying knowledge is learned almost entirely during pretraining, while post-training serves to teach the model correct formatting. Hence, by learning formatting alone, models fine-tuned with as few as 1,000 carefully curated samples can achieve strong alignment capabilities \cite{NEURIPS2023_ac662d74}. Subsequent work focusing on task-specific reasoning finds that while model formatting and stylistic capabilities saturate with as few as 100 samples, underlying reasoning performance continues to improve with more data \cite{raghavendra2024revisitingsuperficialalignmenthypothesis}. Our work extends these findings to the SFT-then-RL paradigm. Minimal, form-only structural supervision is enough to teach the model \textit{form} during SFT, while general reasoning \textit{function} is improved upon during RL. 

Furthermore, our study utilized the OLMo-2-1B base model to systematically sweep a diagnostic grid. Larger language models with more parameters often have stronger reasoning and emergent capabilities \cite{wei2022emergentabilitieslargelanguage}. Consequently, larger base models may require even less SFT structural guidance to initiate effective RL reasoning exploration. We further hypothesize that these findings generalize to other reasoning domains with verifiable outcomes, where minimal SFT establishes required output structure (form) while RL optimizes reasoning strategies (function) through outcome-based rewards, and leave this area as future work.

\subsection{Implications for post-training pipelines}
\label{subsec:synthetic_data}

Modern LLM post-training increasingly relies on synthetic data generation, which can suffer from hallucinations, factual incorrectness, and errors \cite{11080380}. Our earlier finding that SFT accuracy is largely uncorrelated with post-RL accuracy ($r \approx 0$, $p > 0.05$) suggests that filtering synthetic data for strict intermediate-step correctness may be less necessary than enforcing correct outcome verification in RL. Instead, practitioners may tolerate some reasoning errors in synthetic SFT traces or utilize minimal structural SFT, provided structural \textit{form} remains intact. Paired with outcome-reward RL to teach reasoning \textit{function}, this approach may extract strong reasoning performance without spending excessive compute on complex data filtering.

\section*{Limitations}

We address two limitations in our work. First, our evaluation was restricted to the OLMo-2-1B base model and the GSM8K dataset. This allowed us to measure performance from an unsaturated starting point, run computationally extensive sweeps across hundreds of SFT configurations, and utilize a deterministic regex propagation method. However, validating these post-training dynamics on larger models, which possess stronger intrinsic priors, as well as more difficult datasets, remains an area of future work. Second, our error taxonomy covered the majority (93.51\%) of observed numerical reasoning failures, but not all. Future work could explore not only other numerical corruption types, but also further natural language mistakes or self-correction behaviors of LLMs.

\bibliography{custom}

@inproceedings{li-etal-2025-language-models,
    title = "Language Models Can Easily Learn to Reason from Demonstrations",
    author = "Li, Dacheng  and
      Cao, Shiyi  and
      Griggs, Tyler  and
      Liu, Shu  and
      Mo, Xiangxi  and
      Tang, Eric  and
      Hegde, Sumanth  and
      Hakhamaneshi, Kourosh  and
      Patil, Shishir G  and
      Zaharia, Matei  and
      Gonzalez, Joseph E.  and
      Stoica, Ion",
    editor = "Christodoulopoulos, Christos  and
      Chakraborty, Tanmoy  and
      Rose, Carolyn  and
      Peng, Violet",
    booktitle = "Findings of the Association for Computational Linguistics: EMNLP 2025",
    month = nov,
    year = "2025",
    address = "Suzhou, China",
    publisher = "Association for Computational Linguistics",
    url = "https://aclanthology.org/2025.findings-emnlp.866/",
    doi = "10.18653/v1/2025.findings-emnlp.866",
    pages = "15979--15997",
    ISBN = "979-8-89176-335-7"
}

@inproceedings{cho-2024-unveiling,
    title = "Unveiling Imitation Learning: Exploring the impact of Data Falsity to Large Language Model",
    author = "Cho, Hyunsoo",
    editor = "Ku, Lun-Wei  and
      Martins, Andre  and
      Srikumar, Vivek",
    booktitle = "Findings of the Association for Computational Linguistics: ACL 2024",
    month = aug,
    year = "2024",
    address = "Bangkok, Thailand",
    publisher = "Association for Computational Linguistics",
    url = "https://aclanthology.org/2024.findings-acl.6/",
    doi = "10.18653/v1/2024.findings-acl.6",
    pages = "62--73"
}

@misc{olmo20242olmo2furious,
      title={{2 OLMo 2 Furious}},
      author={Team OLMo and Pete Walsh and Luca Soldaini and Dirk Groeneveld and Kyle Lo and Shane Arora and Akshita Bhagia and Yuling Gu and Shengyi Huang and Matt Jordan and Nathan Lambert and Dustin Schwenk and Oyvind Tafjord and Taira Anderson and David Atkinson and Faeze Brahman and Christopher Clark and Pradeep Dasigi and Nouha Dziri and Michal Guerquin and Hamish Ivison and Pang Wei Koh and Jiacheng Liu and Saumya Malik and William Merrill and Lester James V. Miranda and Jacob Morrison and Tyler Murray and Crystal Nam and Valentina Pyatkin and Aman Rangapur and Michael Schmitz and Sam Skjonsberg and David Wadden and Christopher Wilhelm and Michael Wilson and Luke Zettlemoyer and Ali Farhadi and Noah A. Smith and Hannaneh Hajishirzi},
      year={2024},
      eprint={2501.00656},
      archivePrefix={arXiv},
      primaryClass={cs.CL},
      url={https://arxiv.org/abs/2501.00656},
}

@misc{yuan2023largelanguagemodelsperform,
      title={How well do Large Language Models perform in Arithmetic tasks?}, 
      author={Zheng Yuan and Hongyi Yuan and Chuanqi Tan and Wei Wang and Songfang Huang},
      year={2023},
      eprint={2304.02015},
      archivePrefix={arXiv},
      primaryClass={cs.CL},
      url={https://arxiv.org/abs/2304.02015}, 
}

@inproceedings{
matsutani2026rl,
title={{RL} Squeezes, {SFT} Expands: A Comparative Study of Reasoning {LLM}s},
author={Kohsei Matsutani and Shota Takashiro and Gouki Minegishi and Takeshi Kojima and Yusuke Iwasawa and Yutaka Matsuo},
booktitle={The Fourteenth International Conference on Learning Representations},
year={2026},
url={https://openreview.net/forum?id=N2lMNqJsBw}
}

@misc{shao2024deepseekmathpushinglimitsmathematical,
      title={DeepSeekMath: Pushing the Limits of Mathematical Reasoning in Open Language Models}, 
      author={Zhihong Shao and Peiyi Wang and Qihao Zhu and Runxin Xu and Junxiao Song and Xiao Bi and Haowei Zhang and Mingchuan Zhang and Y. K. Li and Y. Wu and Daya Guo},
      year={2024},
      eprint={2402.03300},
      archivePrefix={arXiv},
      primaryClass={cs.CL},
      url={https://arxiv.org/abs/2402.03300}, 
}

@article{Guo_2025,
   title={DeepSeek-R1 incentivizes reasoning in LLMs through reinforcement learning},
   volume={645},
   ISSN={1476-4687},
   url={http://dx.doi.org/10.1038/s41586-025-09422-z},
   DOI={10.1038/s41586-025-09422-z},
   number={8081},
   journal={Nature},
   publisher={Springer Science and Business Media LLC},
   author={Guo, Daya and Yang, Dejian and Zhang, Haowei and Song, Junxiao and Wang, Peiyi and Zhu, Qihao and Xu, Runxin and Zhang, Ruoyu and Ma, Shirong and Bi, Xiao and Zhang, Xiaokang and Yu, Xingkai and Wu, Yu and Wu, Z. F. and Gou, Zhibin and Shao, Zhihong and Li, Zhuoshu and Gao, Ziyi and Liu, Aixin and Xue, Bing and Wang, Bingxuan and Wu, Bochao and Feng, Bei and Lu, Chengda and Zhao, Chenggang and Deng, Chengqi and Ruan, Chong and Dai, Damai and Chen, Deli and Ji, Dongjie and Li, Erhang and Lin, Fangyun and Dai, Fucong and Luo, Fuli and Hao, Guangbo and Chen, Guanting and Li, Guowei and Zhang, H. and Xu, Hanwei and Ding, Honghui and Gao, Huazuo and Qu, Hui and Li, Hui and Guo, Jianzhong and Li, Jiashi and Chen, Jingchang and Yuan, Jingyang and Tu, Jinhao and Qiu, Junjie and Li, Junlong and Cai, J. L. and Ni, Jiaqi and Liang, Jian and Chen, Jin and Dong, Kai and Hu, Kai and You, Kaichao and Gao, Kaige and Guan, Kang and Huang, Kexin and Yu, Kuai and Wang, Lean and Zhang, Lecong and Zhao, Liang and Wang, Litong and Zhang, Liyue and Xu, Lei and Xia, Leyi and Zhang, Mingchuan and Zhang, Minghua and Tang, Minghui and Zhou, Mingxu and Li, Meng and Wang, Miaojun and Li, Mingming and Tian, Ning and Huang, Panpan and Zhang, Peng and Wang, Qiancheng and Chen, Qinyu and Du, Qiushi and Ge, Ruiqi and Zhang, Ruisong and Pan, Ruizhe and Wang, Runji and Chen, R. J. and Jin, R. L. and Chen, Ruyi and Lu, Shanghao and Zhou, Shangyan and Chen, Shanhuang and Ye, Shengfeng and Wang, Shiyu and Yu, Shuiping and Zhou, Shunfeng and Pan, Shuting and Li, S. S. and Zhou, Shuang and Wu, Shaoqing and Yun, Tao and Pei, Tian and Sun, Tianyu and Wang, T. and Zeng, Wangding and Liu, Wen and Liang, Wenfeng and Gao, Wenjun and Yu, Wenqin and Zhang, Wentao and Xiao, W. L. and An, Wei and Liu, Xiaodong and Wang, Xiaohan and Chen, Xiaokang and Nie, Xiaotao and Cheng, Xin and Liu, Xin and Xie, Xin and Liu, Xingchao and Yang, Xinyu and Li, Xinyuan and Su, Xuecheng and Lin, Xuheng and Li, X. Q. and Jin, Xiangyue and Shen, Xiaojin and Chen, Xiaosha and Sun, Xiaowen and Wang, Xiaoxiang and Song, Xinnan and Zhou, Xinyi and Wang, Xianzu and Shan, Xinxia and Li, Y. K. and Wang, Y. Q. and Wei, Y. X. and Zhang, Yang and Xu, Yanhong and Li, Yao and Zhao, Yao and Sun, Yaofeng and Wang, Yaohui and Yu, Yi and Zhang, Yichao and Shi, Yifan and Xiong, Yiliang and He, Ying and Piao, Yishi and Wang, Yisong and Tan, Yixuan and Ma, Yiyang and Liu, Yiyuan and Guo, Yongqiang and Ou, Yuan and Wang, Yuduan and Gong, Yue and Zou, Yuheng and He, Yujia and Xiong, Yunfan and Luo, Yuxiang and You, Yuxiang and Liu, Yuxuan and Zhou, Yuyang and Zhu, Y. X. and Huang, Yanping and Li, Yaohui and Zheng, Yi and Zhu, Yuchen and Ma, Yunxian and Tang, Ying and Zha, Yukun and Yan, Yuting and Ren, Z. Z. and Ren, Zehui and Sha, Zhangli and Fu, Zhe and Xu, Zhean and Xie, Zhenda and Zhang, Zhengyan and Hao, Zhewen and Ma, Zhicheng and Yan, Zhigang and Wu, Zhiyu and Gu, Zihui and Zhu, Zijia and Liu, Zijun and Li, Zilin and Xie, Ziwei and Song, Ziyang and Pan, Zizheng and Huang, Zhen and Xu, Zhipeng and Zhang, Zhongyu and Zhang, Zhen},
   year={2025},
   pages={633–638} 
}

@inproceedings{rajani2025scalpelvshammergrpo,
  title = {Scalpel vs. Hammer: GRPO Amplifies Existing Capabilities, SFT Replaces Them},
  author = {Rajani, Neel and Gema, Aryo Pradipta and Goldfarb-Tarrant, Seraphina and Titov, Ivan},
  year = {2025},
  url = {https://icml.cc/virtual/2025/49627},
  booktitle = {Actionable Interpretability Workshop at ICML 2025},
}

@inproceedings{ICLR2026_2e53c02e,
 author = {Wu, Yongliang and Zhou, Yizhou and Zhou, Ziheng and Peng, Yingzhe and Ye, Xinyu and Hu, Xinting and Zhu, Wenbo and Qi, Lu and Yang, Ming-Hsuan and yang, xu},
 booktitle = {International Conference on Learning Representations},
 editor = {C. Vondrick and B. Hariharan and C. Raffel and L. Pinto and D. Yang and A. Faust},
 pages = {27550--27571},
 title = {On the Generalization of SFT: A Reinforcement Learning Perspective with Reward Rectification},
 url = {https://proceedings.iclr.cc/paper_files/paper/2026/file/2e53c02ea028cbf603f4b6b47fef3d97-Paper-Conference.pdf},
 volume = {2026},
 year = {2026}
}

@inproceedings{wang-etal-2026-learning-staying,
    title = "Learning While Staying Curious: Entropy-Preserving Supervised Fine-Tuning via Adaptive Self-Distillation for Large Reasoning Models",
    author = "Wang, Hao  and
      Gu, Hao  and
      Piao, Hongming  and
      Gong, Kaixiong  and
      Ye, Yuxiao  and
      Yue, Xiangyu  and
      Han, Sirui  and
      Guo, Yike  and
      Wu, Dapeng",
    editor = "Liakata, Maria  and
      Moreira, Viviane P.  and
      Zhang, Jiajun  and
      Jurgens, David",
    booktitle = "Proceedings of the 64th Annual Meeting of the {A}ssociation for {C}omputational {L}inguistics (Volume 1: Long Papers)",
    month = jul,
    year = "2026",
    address = "San Diego, California, United States",
    publisher = "Association for Computational Linguistics",
    url = "https://aclanthology.org/2026.acl-long.617/",
    doi = "10.18653/v1/2026.acl-long.617",
    pages = "13567--13581",
    ISBN = "979-8-89176-390-6"
}

@inproceedings{ICLR2026_5d6ae8ba,
 author = {Kang, Feiyang and Kuchnik, Michael and Padthe, Karthik and Vlastelica, Marin and Jia, Ruoxi and Wu, Carole-Jean and Ardalani, Newsha},
 booktitle = {International Conference on Learning Representations},
 editor = {C. Vondrick and B. Hariharan and C. Raffel and L. Pinto and D. Yang and A. Faust},
 pages = {56876--56918},
 title = {Quagmires in SFT-RL Post-Training: When High SFT Scores Mislead and What to Use Instead},
 url = {https://proceedings.iclr.cc/paper_files/paper/2026/file/5d6ae8ba43ecb378030753c4408ef9bd-Paper-Conference.pdf},
 volume = {2026},
 year = {2026}
}

@misc{chu2025sftmemorizesrlgeneralizes,
      title={SFT Memorizes, RL Generalizes: A Comparative Study of Foundation Model Post-training}, 
      author={Tianzhe Chu and Yuexiang Zhai and Jihan Yang and Shengbang Tong and Saining Xie and Dale Schuurmans and Quoc V. Le and Sergey Levine and Yi Ma},
      year={2025},
      eprint={2501.17161},
      archivePrefix={arXiv},
      primaryClass={cs.AI},
      url={https://arxiv.org/abs/2501.17161}, 
}

@misc{cobbe2021trainingverifierssolvemath,
      title={Training Verifiers to Solve Math Word Problems}, 
      author={Karl Cobbe and Vineet Kosaraju and Mohammad Bavarian and Mark Chen and Heewoo Jun and Lukasz Kaiser and Matthias Plappert and Jerry Tworek and Jacob Hilton and Reiichiro Nakano and Christopher Hesse and John Schulman},
      year={2021},
      eprint={2110.14168},
      archivePrefix={arXiv},
      primaryClass={cs.LG},
      url={https://arxiv.org/abs/2110.14168}, 
}

@misc{dragoi2025passkbreadthdepthmetricsreasoning,
      title={Beyond Pass@k: Breadth-Depth Metrics for Reasoning Boundaries}, 
      author={Marius Dragoi and Ioana Pintilie and Florin Gogianu and Florin Brad},
      year={2025},
      eprint={2510.08325},
      archivePrefix={arXiv},
      primaryClass={cs.AI},
      url={https://arxiv.org/abs/2510.08325}, 
}

@inproceedings{NEURIPS2023_ac662d74,
 author = {Zhou, Chunting and Liu, Pengfei and Xu, Puxin and Iyer, Srinivasan and Sun, Jiao and Mao, Yuning and Ma, Xuezhe and Efrat, Avia and Yu, Ping and YU, LILI and Zhang, Susan and Ghosh, Gargi and Lewis, Mike and Zettlemoyer, Luke and Levy, Omer},
 booktitle = {Advances in Neural Information Processing Systems},
 doi = {10.52202/075280-2400},
 editor = {A. Oh and T. Naumann and A. Globerson and K. Saenko and M. Hardt and S. Levine},
 pages = {55006--55021},
 publisher = {Curran Associates, Inc.},
 title = {LIMA: Less Is More for Alignment},
 url = {https://proceedings.neurips.cc/paper_files/paper/2023/file/ac662d74829e4407ce1d126477f4a03a-Paper-Conference.pdf},
 volume = {36},
 year = {2023}
}

@article{Zhao2026,
    author = {Zhao, Wayne Xin and Zhou, Kun and Li, Junyi and Tang, Tianyi and Dong, Zican and Hou, Yupeng and Zhang, Beichen and Min, Yingqian and Zhang, Junjie and Liu, Peiyu and Wang, Xiaolei and Du, Yifan and Yang, Chen and Chen, Yushuo and Chen, Zhipeng and Jiang, Jinhao and Ren, Ruiyang and Li, Yifan and Tang, Xinyu and Liu, Zikang and Hu, Yiwen and Nie, Jian-Yun and Wen, Ji-Rong},
    title = {A Survey of Large Language Models},
    journal = {Frontiers of Computer Science},
    year = {2026},
    volume = {20},
    number = {12},
    pages = {2012627},
    doi = {10.1007/s11704-026-60308-3},
    url = {https://doi.org/10.1007/s11704-026-60308-3}
}

@inproceedings{NIPS2014_5a18e133,
 author = {Sutskever, Ilya and Vinyals, Oriol and Le, Quoc V.},
 booktitle = {Advances in Neural Information Processing Systems},
 editor = {Z. Ghahramani and M. Welling and C. Cortes and N. Lawrence and K. Weinberger},
 pages = {},
 publisher = {Curran Associates, Inc.},
 title = {Sequence to Sequence Learning with Neural Networks},
 url = {https://proceedings.neurips.cc/paper_files/paper/2014/file/5a18e133cbf9f257297f410bb7eca942-Paper.pdf},
 volume = {27},
 year = {2014}
}

@inproceedings{NIPS2017_3f5ee243,
 author = {Vaswani, Ashish and Shazeer, Noam and Parmar, Niki and Uszkoreit, Jakob and Jones, Llion and Gomez, Aidan N and Kaiser, Lukasz and Polosukhin, Illia},
 booktitle = {Advances in Neural Information Processing Systems},
 editor = {I. Guyon and U. Von Luxburg and S. Bengio and H. Wallach and R. Fergus and S. Vishwanathan and R. Garnett},
 pages = {},
 publisher = {Curran Associates, Inc.},
 title = {Attention is All you Need},
 url = {https://proceedings.neurips.cc/paper_files/paper/2017/file/3f5ee243547dee91fbd053c1c4a845aa-Paper.pdf},
 volume = {30},
 year = {2017}
}

@inproceedings{magister-etal-2023-teaching,
    title = "Teaching Small Language Models to Reason",
    author = "Magister, Lucie Charlotte  and
      Mallinson, Jonathan  and
      Adamek, Jakub  and
      Malmi, Eric  and
      Severyn, Aliaksei",
    editor = "Rogers, Anna  and
      Boyd-Graber, Jordan  and
      Okazaki, Naoaki",
    booktitle = "Proceedings of the 61st Annual Meeting of the Association for Computational Linguistics (Volume 2: Short Papers)",
    month = jul,
    year = "2023",
    address = "Toronto, Canada",
    publisher = "Association for Computational Linguistics",
    url = "https://aclanthology.org/2023.acl-short.151/",
    doi = "10.18653/v1/2023.acl-short.151",
    pages = "1773--1781"
}

@article{JMLR:v25:23-0870,
  author  = {Hyung Won Chung and Le Hou and Shayne Longpre and Barret Zoph and Yi Tay and William Fedus and Yunxuan Li and Xuezhi Wang and Mostafa Dehghani and Siddhartha Brahma and Albert Webson and Shixiang Shane Gu and Zhuyun Dai and Mirac Suzgun and Xinyun Chen and Aakanksha Chowdhery and Alex Castro-Ros and Marie Pellat and Kevin Robinson and Dasha Valter and Sharan Narang and Gaurav Mishra and Adams Yu and Vincent Zhao and Yanping Huang and Andrew Dai and Hongkun Yu and Slav Petrov and Ed H. Chi and Jeff Dean and Jacob Devlin and Adam Roberts and Denny Zhou and Quoc V. Le and Jason Wei},
  title   = {Scaling Instruction-Finetuned Language Models},
  journal = {Journal of Machine Learning Research},
  year    = {2024},
  volume  = {25},
  number  = {70},
  pages   = {1--53},
  url     = {http://jmlr.org/papers/v25/23-0870.html}
}

@inproceedings{NEURIPS2022_b1efde53,
 author = {Ouyang, Long and Wu, Jeffrey and Jiang, Xu and Almeida, Diogo and Wainwright, Carroll and Mishkin, Pamela and Zhang, Chong and Agarwal, Sandhini and Slama, Katarina and Ray, Alex and Schulman, John and Hilton, Jacob and Kelton, Fraser and Miller, Luke and Simens, Maddie and Askell, Amanda and Welinder, Peter and Christiano, Paul F and Leike, Jan and Lowe, Ryan},
 booktitle = {Advances in Neural Information Processing Systems},
 doi = {10.52202/068431-2011},
 editor = {S. Koyejo and S. Mohamed and A. Agarwal and D. Belgrave and K. Cho and A. Oh},
 pages = {27730--27744},
 publisher = {Curran Associates, Inc.},
 title = {Training language models to follow instructions with human feedback},
 url = {https://proceedings.neurips.cc/paper_files/paper/2022/file/b1efde53be364a73914f58805a001731-Paper-Conference.pdf},
 volume = {35},
 year = {2022}
}

@misc{wen2025reinforcementlearningverifiablerewards,
      title={Reinforcement Learning with Verifiable Rewards Implicitly Incentivizes Correct Reasoning in Base LLMs}, 
      author={Xumeng Wen and Zihan Liu and Shun Zheng and Shengyu Ye and Zhirong Wu and Yang Wang and Zhijian Xu and Xiao Liang and Junjie Li and Ziming Miao and Jiang Bian and Mao Yang},
      year={2025},
      eprint={2506.14245},
      archivePrefix={arXiv},
      primaryClass={cs.AI},
      url={https://arxiv.org/abs/2506.14245}, 
}

@misc{wei2022emergentabilitieslargelanguage,
      title={Emergent Abilities of Large Language Models}, 
      author={Jason Wei and Yi Tay and Rishi Bommasani and Colin Raffel and Barret Zoph and Sebastian Borgeaud and Dani Yogatama and Maarten Bosma and Denny Zhou and Donald Metzler and Ed H. Chi and Tatsunori Hashimoto and Oriol Vinyals and Percy Liang and Jeff Dean and William Fedus},
      year={2022},
      eprint={2206.07682},
      archivePrefix={arXiv},
      primaryClass={cs.CL},
      url={https://arxiv.org/abs/2206.07682}, 
}

@ARTICLE{11080380,
  author={Nadǎş, Mihai and Dioşan, Laura and Tomescu, Andreea},
  journal={IEEE Access}, 
  title={Synthetic Data Generation Using Large Language Models: Advances in Text and Code}, 
  year={2025},
  volume={13},
  number={},
  pages={134615-134633},
  doi={10.1109/ACCESS.2025.3589503}}

@misc{raghavendra2024revisitingsuperficialalignmenthypothesis,
      title={Revisiting the Superficial Alignment Hypothesis}, 
      author={Mohit Raghavendra and Vaskar Nath and Sean Hendryx},
      year={2024},
      eprint={2410.03717},
      archivePrefix={arXiv},
      primaryClass={cs.CL},
      url={https://arxiv.org/abs/2410.03717}, 
}

@misc{zhang2026onpolicyrlmeetsoffpolicy,
      title={On-Policy RL Meets Off-Policy Experts: Harmonizing Supervised Fine-Tuning and Reinforcement Learning via Dynamic Weighting}, 
      author={Wenhao Zhang and Yuexiang Xie and Yuchang Sun and Yanxi Chen and Guoyin Wang and Yaliang Li and Bolin Ding and Jingren Zhou},
      year={2026},
      eprint={2508.11408},
      archivePrefix={arXiv},
      primaryClass={cs.LG},
      url={https://arxiv.org/abs/2508.11408}, 
}

\appendix

\section{Corrupter Details}
\label{appendix:corrupter_details}

\subsection{Pipeline Mechanics}

\textbf{Equation selection.} Regex first finds an equation to corrupt; this equation must have an equal sign and at least one operator (e.g. equations such as \texttt{\textless\textless14=14\textgreater\textgreater} are ignored). Our experiments include corrupting both the \textit{first} and \textit{last} regex-extracted equations.

\textbf{Error propagation.} Once an equation is selected and corrupted, the new value representing the error in the RHS is propagated throughout the rest of the trace, while all other original text is preserved. If the left-hand side (LHS) was mutated, the new string is parsed and evaluated using a math string evaluator to determine the new right-hand side (RHS). Division-by-zero errors are caught and replaced with a large constant ($2.92 \times 10^{9}$) to prevent pipeline crashes. To propagate, the new RHS value overwrites the old one anywhere it appears downstream. Subsequent equations are recalculated and follow this logic with their new RHS values. Natural language words such as \texttt{plus, minus, times, divided by, makes, leaving} are treated as operations. The final answer after \#\#\#\# is rewritten with the modified value after all the relevant downstream calculations. 

\subsection{Error Taxonomy}

\textbf{Hallucination error.} 
The corrupter deterministically chooses an operand on the LHS of the extracted equation and replaces it with a value sampled from a uniform distribution bounded between 1 and 1000. If the operand is a decimal then the sampled value matches its precision. 

\textbf{Operator error.} The corrupter identifies arithmetic operators in the LHS and swaps addition with subtraction, and multiplication with division. 

\textbf{Operand error.} The corrupter extracts both English-word numbers and digits from the text of the \textit{question} and converts them into a list of usable floats. If the LHS of the extracted equation in the trace contains two operands with a non-commutative operator (subtraction or division), the positions of the two operands are simply inverted. Otherwise, one operand in the LHS is randomly replaced with one value from the prior list of usable floats. 

\textbf{Add step error.} The corrupter once again extracts English-word numbers and digits from the text of the question, and examines existing operators in the LHS. Then a cartesian product of existing operators and extracted question numbers is built. Combinations that result in a division by zero, or multiplication or division by 1, are dropped, and a remaining valid combination is randomly selected and inserted into the LHS of the trace.  

\textbf{Forget step error.} The corrupter deletes an operand and relevant operator from the LHS of the extracted equation. Specifically, this operand is \textit{not} the first operand, in order to mimic the forgetting of a step. Equations with three or more operands retain operators, while simple binary equations turn into a simple identity (e.g. \texttt{\textless\textless14+2=16\textgreater\textgreater} becomes \texttt{\textless\textless14=14\textgreater\textgreater}).

\textbf{Calculation error.} In the extracted equation, the corrupter checks that the right-hand side (RHS) is non-zero, otherwise the next equation is selected (or, the second-to-last equation if the last regex equation is being corrupted). A multiplicative ratio is applied to the RHS directly based on a statistical distribution of analyzed calculation errors. 

More precisely, this is a Weibull distribution ($c=0.578, \lambda = 0.069$) fitted to observed calculation errors in the diagnostic GRPO run. To determine if the model is underestimating or overestimating the true value, a negative direction is assigned with probability $\frac{52}{91}$. The signed log-ratio is exponentiated into a multiplicative ratio. Rare sign-inversion errors are accounted for with a final negative probability $\frac{1}{91}$.

\section{Trace Structure and Evaluation Protocol}
\label{appendix:evaluation_details}
All traces are formatted with specific delimiters for formatting. The original GSM8K and form-only questions are wrapped by  \texttt{\textless\textless question\textgreater\textgreater}  and \texttt{\textless\textless/question\textgreater\textgreater} tokens, the reasoning chains by \texttt{\textless\textless think\textgreater\textgreater}  and \texttt{\textless\textless/think\textgreater\textgreater} tokens, and the final answer is prepended with a {\#\#\#\#} delimiter. To evaluate correctness,  we use a strict evaluator enforcing exact pattern matching on terminal \texttt{\#\#\#\#} answer blocks on the GSM8K test set.

\section{Computational Experimental Details}
\label{appendix:hyperparameters}
 SFT parameters (learning rate = $2\times10^{-5}$, 4 epochs, batch size = 4) were selected by sweeping learning rates from $10^{-5}$ to $10^{-3}$ and training durations from 1 to 10 epochs. GRPO parameters (learning rate = $3\times10^{-7}$, constant learning-rate schedule with 3\% warmup, $\beta$=0.01, batch size = 4) were selected through sweeps over the first 10,000 training steps. Training was continued until performance saturated. 8 rollouts per optimization were used, with a maximum completion length of 512 tokens, covering more than 90\% of generated responses. For regex propagation, we use version 2.2.1 of the Python regular expression package. 
 
 Due to the iterative nature of our project over several months, exact hardware logs for early sweeps were not preserved. For GPU hours, we provide an estimate of 1.92 GPU hours per SFT GSM8K configuration, $0.002- 0.0071$ GPU hours per form-only SFT configuration, and 2.85 GPU hours per GRPO run. All models were trained in bfloat 16 precision and on NVIDIA RTX PRO 6000 Blackwell GPUs.

\section{SFT-Only Corruption Analysis}
\label{appendix:sft_corruption_effects}
Here, we analyze how specific training data corruption factors influenced the SFT degradation seen in our results. 

\begin{figure}[t]
    \centering
    \includegraphics[width=\columnwidth]{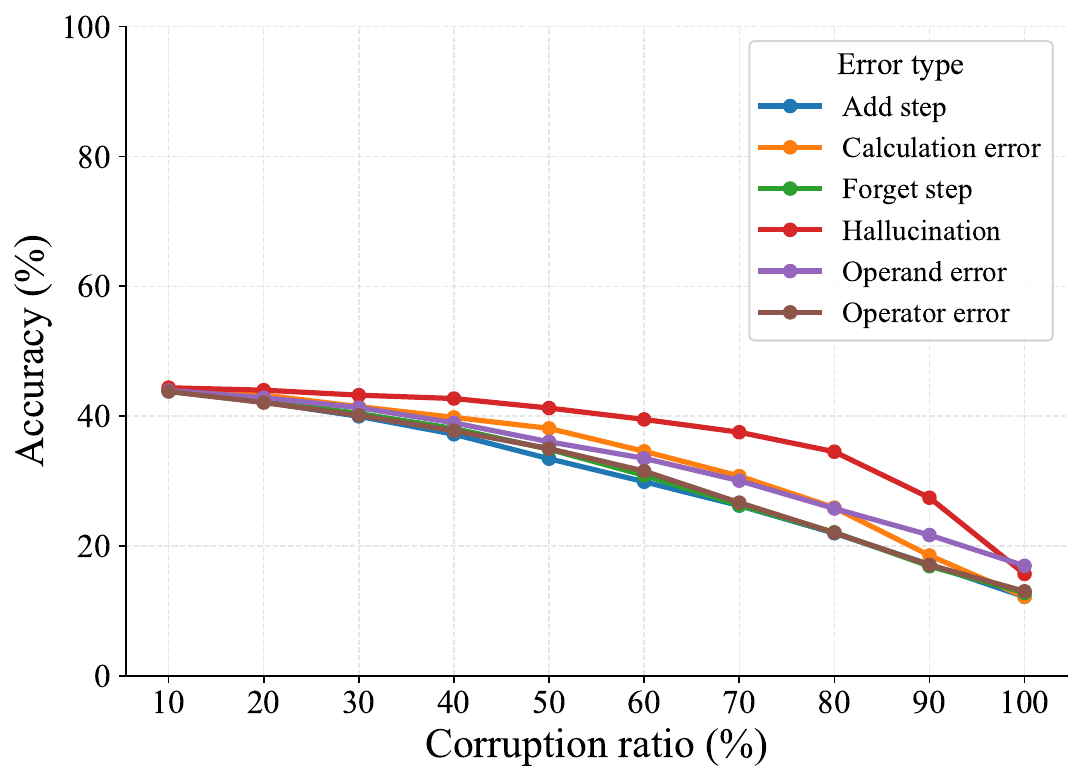}
    \caption{All corruption types exhibit a downwards trend in accuracy as corruption ratio increases. Results are averaged across all SFT configurations at pass@$2$.}
    \label{fig:corruption_type}
\end{figure}

(1) We find that the \textit{hallucination} error type resulted in slower degradation than other error types, but all six error types still showed a consistent downward trend in SFT model performance as the fraction of corrupted data increased (Figure \ref{fig:corruption_type}). 

\begin{figure}[t]
    \centering
    \includegraphics[width=\columnwidth]{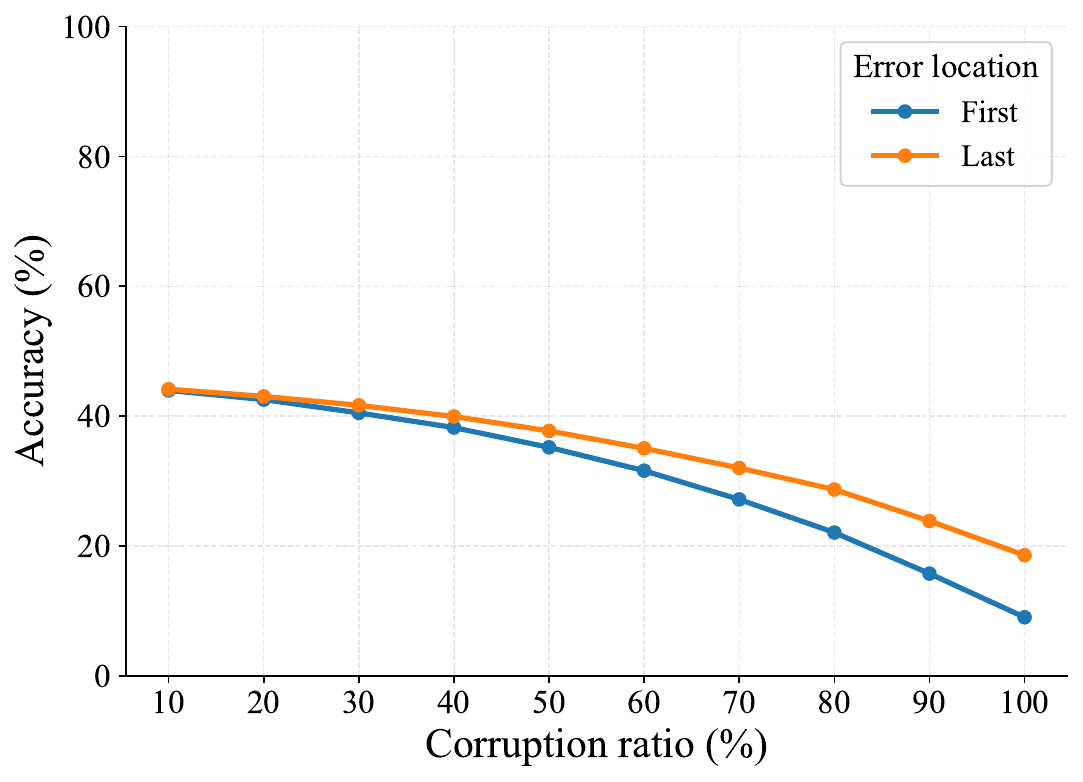}
    \caption{Corrupting the first equation leads to more degradation than corrupting the last equation. Results are averaged across all SFT configurations at pass@$2$.}
    \label{fig:location_type}
\end{figure}

(2) \textit{Error location} impacts results; corrupting the first equation in the trace results in more performance degradation than corrupting the last equation (Figure \ref{fig:location_type}). 

\begin{figure}[t]
    \centering
    \includegraphics[width=\columnwidth]{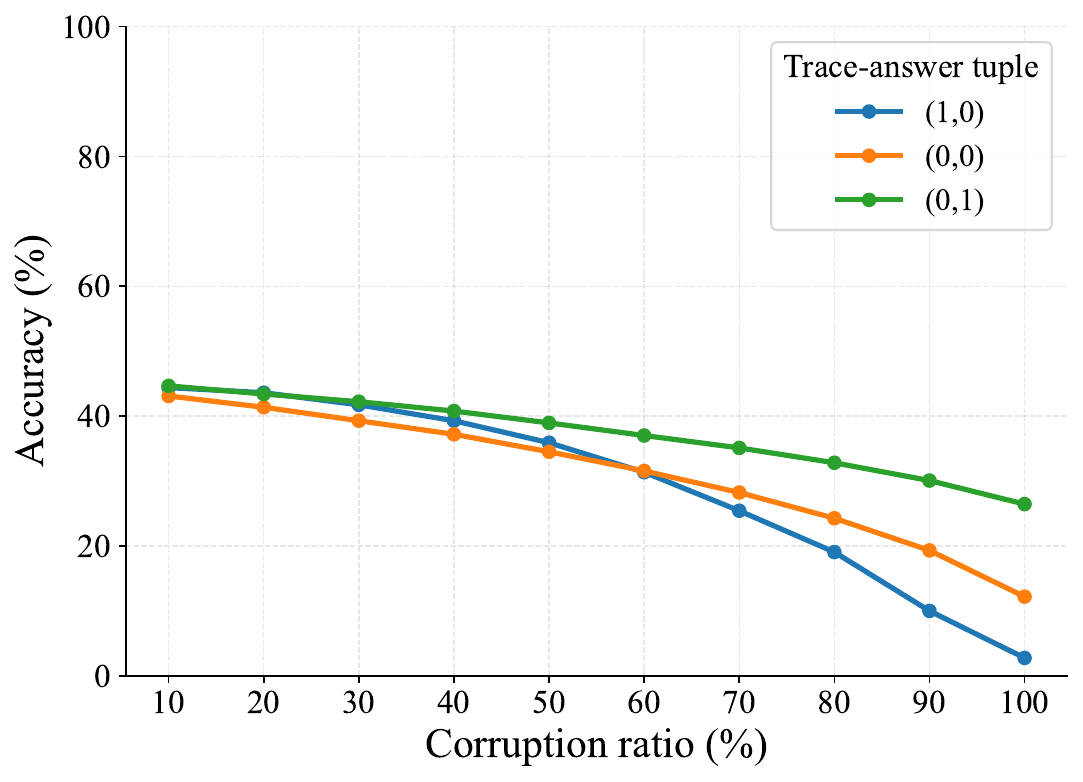}
    \caption{The (1,0) trace-answer configuration leads to the most degradation as the corruption ratio increased, followed by the (0,0) and (0,1) tuples. Results are averaged across all SFT configurations at pass@$2$.}
    \label{fig:tuple_type}
\end{figure}

(3) The (1,0) trace-answer configuration experiences the most degradation, followed by the (0,0) configuration and finally the (0,1) configuration (Figure \ref{fig:tuple_type}). 

\section{Other Pass@\texorpdfstring{$k$}{k} Results}
\label{appendix:other_pass_@_k_results}

Our experimental results hold across differing pass@$k$ values. 

\begin{figure}[t]
    \centering
    \includegraphics[width=\columnwidth]{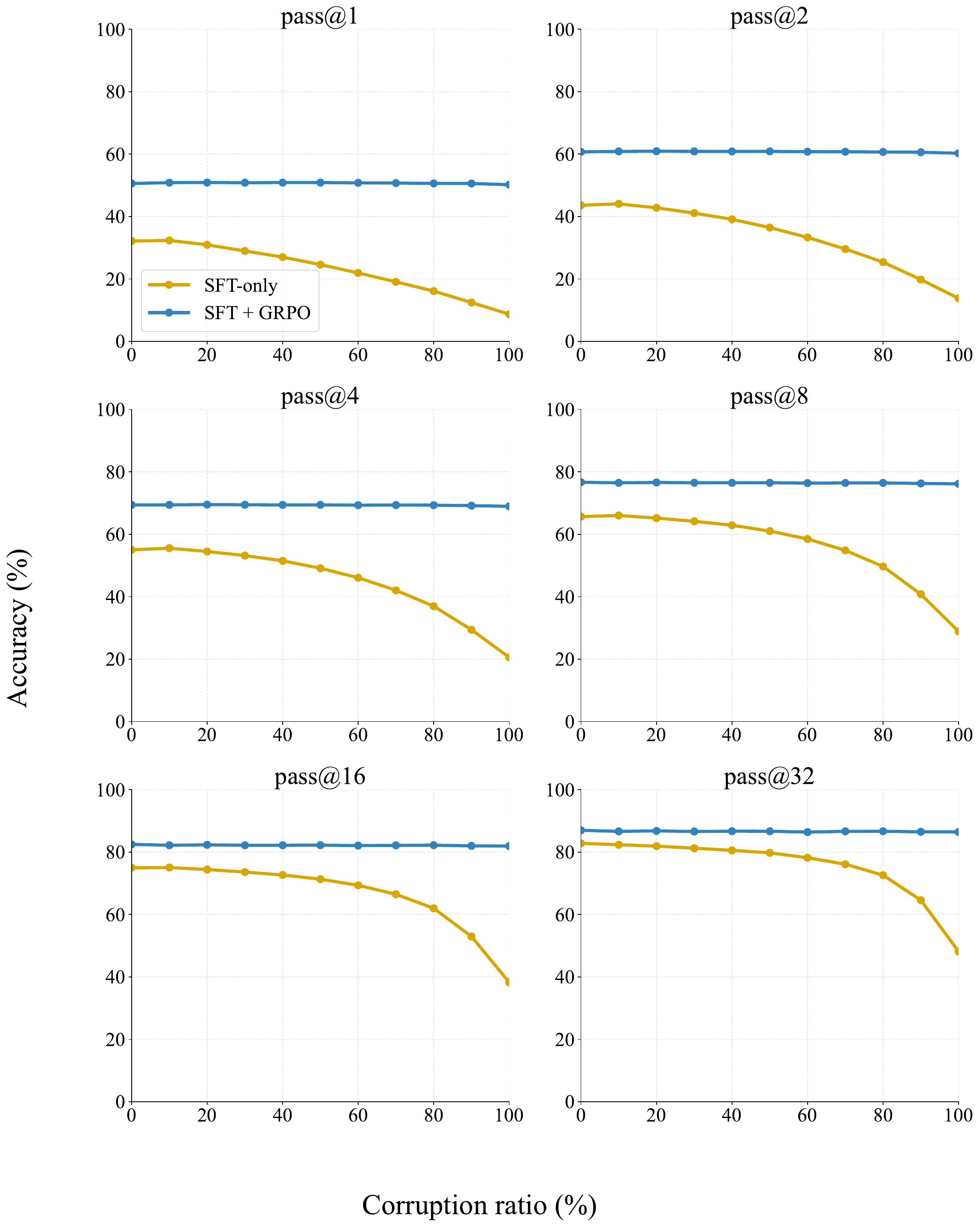}
    \caption{As corruption ratio increases, aggregated SFT-only performance deteriorates, but aggregated GRPO post-training corrects this behavior across all pass@$k$ values.}
    \label{fig:sft_vs_grpo_all_k}
\end{figure}

(1)  GRPO applied to corrupted SFT recovers performance to near-gold baseline levels at all corruption ratios (Figure \ref{fig:sft_vs_grpo_all_k}). 

\begin{figure}[t]
    \centering
    \includegraphics[width=\columnwidth]{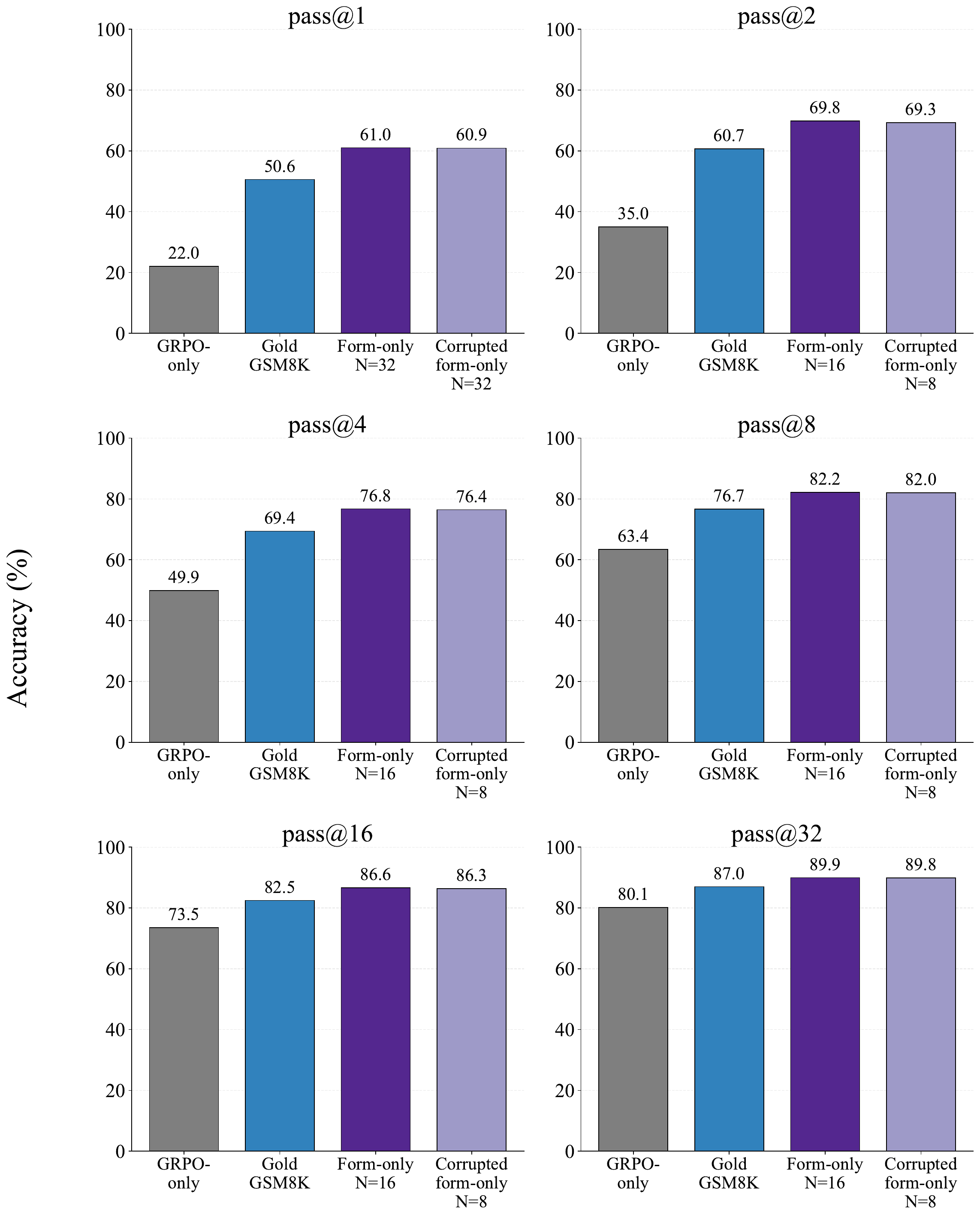}
    \caption{SFT on few form-only samples, whether corrupted or regular, outperforms both gold GSM8K initialization and GRPO-only baselines across all pass@$k$ values.}
    \label{fig:initializations_all_k}
\end{figure}

(2) Form-only initializations outperform standard SFT on gold GSM8K traces, as well as each GRPO-only baseline (Figure \ref{fig:initializations_all_k}). 

As k increases, performance gaps between each initialization method decrease. For example, pass@$32$ results in a high GRPO-only baseline accuracy of 80.1\%, and resulting gains of other initializations are less than 10 percentage points higher. However, pass$@$1 displayed a 39 percentage point jump from the GRPO-only baseline to form-only initializations. Furthermore, while Figure \ref{fig:initializations_all_k} only displays gold GSM8K initializations, no GSM8K initialization outperforms all regular or corrupted form-only configurations at any pass@$k$. Across every pass@$k$, at least one form-only initialization achieves higher accuracy than the best GSM8K initialization. 

\section{AI Assistants in Research}
AI assistants were used for language assistance and polishing of the writing, as well as to debug author code as needed. All ideas, experimental designs,  and scientific arguments are the authors' original work. Furthermore, all code was written, reviewed, and validated by the authors. All AI Assistant usage conformed with the ACL publications ethics policy. 

\end{document}